\PassOptionsToPackage{hypertexnames=false}{hyperref}
\documentclass[pdflatex,sn-mathphys-num,iicol]{sn-jnl}
\usepackage{graphicx}
\usepackage{multirow}
\usepackage{amsmath,amssymb,amsfonts}
\usepackage{amsthm}
\usepackage[title]{appendix}
\usepackage{xcolor}
\usepackage{textcomp}
\usepackage{manyfoot}
\usepackage{booktabs}
\usepackage{stfloats}
\usepackage{xurl}
\newcounter{supplementsection}
\newcommand{\supplementsection}[1]{\stepcounter{supplementsection}\section*{S1-\arabic{supplementsection}. #1}}
\makeatletter
\providecommand{\burl}[1]{\url{#1}}
\DeclareRobustCommand{\toolname}{Backend--Frontend Converter}
\DeclareRobustCommand{\papertitle}{Backend--Frontend Converter: Component-Based Conversion of .NET Computational Models to TypeScript}
\providecommand{\orcidID}[1]{\g@addto@macro\artauthors{\textsuperscript{\smash{[#1]}}}}
\makeatother

\hypersetup{pdftitle={Backend-Frontend Converter: Component-Based Conversion of .NET Computational Models to TypeScript},pdfauthor={Peter Ivankov and Max Shestov}}
\renewcommand{\figurename}{Fig.}
\begin{document}
\title[\toolname{}]{\papertitle{}}
\author[1]{\fnm{Petr} \sur{Ivankov}}\orcidID{0000-0002-2977-8567}\email{monster.ivankov@gmail.com}
\author*[2]{\fnm{Max} \sur{Shestov}}\orcidID{0000-0002-6759-1744}\email{mshestov@gatech.edu}
\affil[1]{\orgname{Independent Researcher}, \city{Moscow}, \country{Russia}}
\affil[2]{\orgname{Georgia Institute of Technology}, \city{Atlanta}, \state{GA}, \country{United States}}
\abstract{\textbf{Purpose} Computational models developed in .NET often need to run locally within web applications. We present Backend--Frontend Converter, an architecture for generating TypeScript implementations of component-based models from BP\_Simulator.
\par\noindent\textbf{Methods} The converter maps components, connections, aliases, and formula trees to model-specific TypeScript code. Reusable runtime libraries provide numerical integration, data access, and state handling. We evaluate stored generated implementations through four orbital scenarios and a recorded trading-strategy comparison. Orbital calculations execute in a browser and are paired with C\# results by crossing event and timestamp; trading outputs are aligned by date. Analytic scheduler tests and deliberately perturbed outputs assess the validation procedure.
\par\noindent\textbf{Results} Each orbital scenario produces 91 paired events with identical timestamps. Position and velocity differences remain below \(10^{-6}\)~km and \(10^{-9}\)~km/s. All 20 analytic scheduler checks pass, and the comparator detects each of three prescribed state and time perturbations. The trading comparison contains 501 aligned daily records with matching indicator values, positions, and order events.
\par\noindent\textbf{Conclusion} The evaluated implementations preserve the compared outputs across the C\#/TypeScript boundary. Component-specific generation and reusable execution libraries provide a practical route from supported .NET computational models to TypeScript applications.}

\keywords{Model-driven engineering, Code generation, Component-based software engineering, Model migration, TypeScript, Computational models}

\maketitle
\section{Introduction}

\begin{table*}[!b]
\caption{Migration contract for executable computational models. The contract defines which source-side model roles must have target-side TypeScript counterparts and how each role is checked during migration.}
\label{tab:migration-contract}
\centering
\small
\begin{tabular}{@{}p{0.14\linewidth}p{0.23\linewidth}p{0.28\linewidth}p{0.25\linewidth}@{}}
\hline
Model role & Source-side requirement & Target-side TypeScript requirement & Validation check \\
\hline
Component & Explicit runtime component & TypeScript component class or interface & Presence and behavior \\
Alias & Stable semantic name & Same alias in TypeScript & State initialization and output access \\
Measurement & Typed value access & Measurement map or interface & Value comparison \\
Processor & Known execution semantics & TypeScript processor implementation & Trajectory or sequence comparison \\
Domain plugin & Known domain behavior & TypeScript plugin implementation & Domain scenario tests \\
Data consumer & Output collection role & TypeScript consumer & Output alignment \\
\hline
\end{tabular}
\end{table*}

Computational models developed in .NET often combine numerical algorithms, domain-specific components, and execution logic tied to a backend or desktop environment. Running these models locally in a web application supports interactive feedback and reduces the need for server communication. A manual rewrite, however, duplicates domain logic and requires developers to maintain corresponding implementations.

This paper presents \toolname{}, a converter from .NET computational component models to TypeScript. The input is a graph of components and connections in the BP\_Simulator modeling environment. Component mappings translate its configuration, formula trees, value bindings, and connections into executable TypeScript code. Runtime libraries provide the numerical processors and domain behavior used by the generated model.

The central engineering problem is to preserve how a computational model uses its components: formulas must read the intended measurements, aliases must identify the same state variables, and processors must advance and collect that state in the intended order. The migration contract in Table~\ref{tab:migration-contract} connects these model roles to generation rules and runtime responsibilities.

The paper contributes an implementation of this contract for computational component graphs and an evaluation spanning continuous and discrete calculations. Orbital forecasting combines numerical integration with event-based trajectory collection; a trading strategy combines filters, recursive state, and order generation. Comparisons with the corresponding C\# implementations use explicit alignment rules, numerical tolerances, and reproducible comparison scripts to assess output agreement in both cases.

\section{Background and Motivation}

Model-driven architecture separates a model from the platform-specific artifacts that realize it~\cite{object_management_group_mda_2003,mde}. In the present setting, an executable computational model supplies the components and connections used for code generation. Reusable components encapsulate defined execution roles, following component-based software engineering~\cite{szyperski_component_2002}. This use of models to generate application code also relates to low-code platforms~\cite{sahay_supporting_2020}.

Several deployment choices are possible. An application programming interface (API) keeps computation on the server, while WebAssembly provides a route to executing compiled code in the browser~\cite{haas_2017}. A TypeScript implementation lets the computational model use the surrounding web application's libraries and runtime directly.

Practical language conversion requires mappings between source and target constructs~\cite{terekhov2000}. Computational components provide a useful basis for these mappings because their interfaces expose both the values they exchange and their execution roles. The converter uses this information to connect generated formulas and configuration to the TypeScript runtime.

\section{Component-Based Migration Contract}

\toolname{} operates on a computational model represented by .NET component objects and interfaces. Named components, connections, aliases, measurements, execution processors, output consumers, and domain extensions describe its structure and behavior. C\# and TypeScript provide the platform-specific implementations of these roles.

The source modeling environment is \texttt{BP\_Simulator}. Its Windows Forms application, \texttt{BP\_Simulator.Light}, is titled \emph{Business Process Simulation}. The editor, generators, runtime libraries, and example models are obtainable as source from the MIT-licensed repository identified in the Code Availability statement. The editor saves .NET component graphs in its \texttt{.business\_analisys} format; the generator traverses the graph through \texttt{IComponentCollection} and emits supported components and connections.

The migration contract specifies the correspondences required for conversion (Table~\ref{tab:migration-contract}). Each supported component has a generation rule and the runtime support used by its generated code. Aliases retain the names used to initialize and access values, measurement bindings retain their value sources, and connections retain their endpoints. Processors and domain components implement the corresponding execution behavior in TypeScript.

Generation constructs and configures the component graph. For formula components, it also translates formula trees into executable expressions and binds their inputs to measurements. Developers supply the component mappings and runtime implementations; the generator applies them to each model.

\subsection{Abstraction Levels and Conversion Boundary}

Three levels organize the conversion. The model level contains component instances, parameters, formulas, and connection endpoints. The generation level maps these elements to target classes, calculations, and construction code. The runtime level supplies component behavior and execution semantics shared by the generated models. The implementation of these levels and the extension mechanism are described in the next section.

\subsection{Reference-Based Validation Protocol}

The evaluation combines structural traceability with comparisons of executed model outputs. Numerical values are compared within stated tolerances, while discrete states and events are compared exactly. Orbital outputs are paired by crossing-event index, with exact timestamp agreement; trading outputs are aligned by date and use a shared frozen input. The case studies report the resulting agreement for the evaluated scenarios.

\subsection{Use of Artificial Intelligence Tools}
ChatGPT (OpenAI) assisted manuscript preparation, code review, and development of the validation scripts. The authors reviewed and verified its outputs and take full responsibility for the manuscript.

\section{Tool Architecture}
\begin{figure*}[t]
\centering
\includegraphics[width=0.86\textwidth]{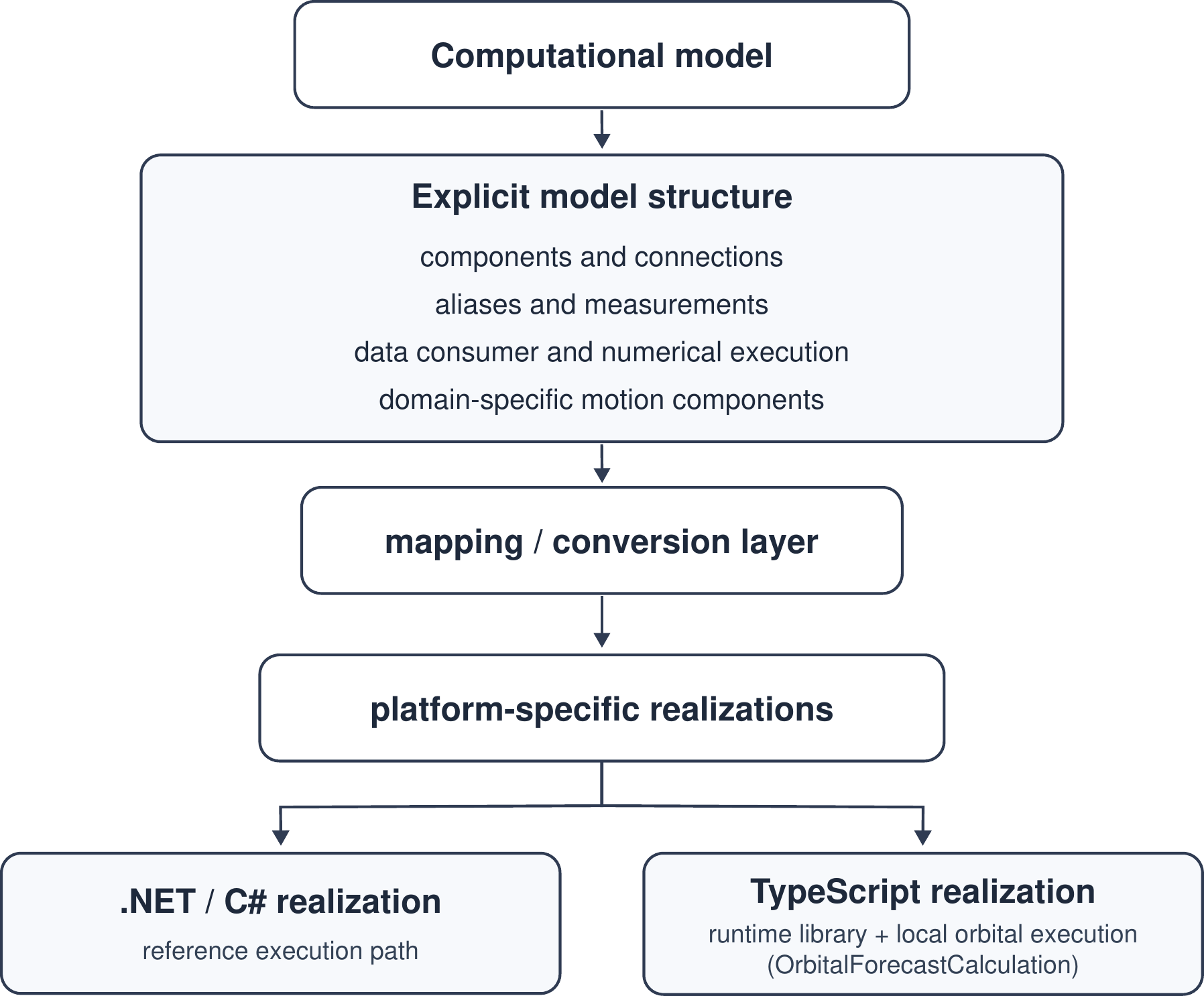}
\caption{Component-based mapping used by \toolname{}. The computational model is represented through explicit runtime structures and realized either by the .NET/C\# execution path or by the TypeScript runtime used in the orbital forecasting implementation}
\label{figure_model_driven_component_mapping}
\end{figure*}

\toolname{} has a generation layer, a target runtime, and a web application for execution and comparison. Figure~\ref{figure_model_driven_component_mapping} shows the relationship between the model and its platform implementations.

The .NET/C\# implementation supplies the source model and the reference execution used for output comparison.

The TypeScript code generator \texttt{DesktopCodeCreator} accepts an \texttt{IComponentCollection}. It enumerates component objects and connecting arrows, delegates their code generation to registered \texttt{IClassCodeCreator} implementations, and emits a model class derived from the TypeScript \texttt{Desktop}. Its constructor creates the generated component instances; its \texttt{finish()} method assigns arrow sources and targets and performs the required post-connection initialization. Formula generators emit calculation methods and measurement bindings from formula trees.

To support a new component family, a developer adds its \texttt{IClassCodeCreator} mapping and the required target runtime behavior. The mapping specifies how the component's parameters and value bindings are emitted. The trading components illustrate this mechanism: a source \texttt{DataQuery} produces a subclass of \texttt{TradingDataQuery} whose constructor copies the configured date interval, period, and symbol. An \texttt{Order} component maps to \texttt{TradingOrder} and supplies the names of its position, price, and date measurements. The runtime classes implement historical-data access and order behavior; the generated subclasses configure them for each model.

The generated code uses the TypeScript runtime described in the next section, including an ordinary differential equation (ODE) processor. The repository provides a worked conversion procedure for the FeedBackFormula model, identifying its source model, generator entry point, generated TypeScript, and runtime dependencies.

The React client can execute the TypeScript orbital calculation locally or call the ASP.NET Core reference endpoint and display the results for comparison.

Table~\ref{tab:traceability} relates the source and target elements used in orbital forecasting.

\begin{table*}[t]
\centering
\caption{Traceability between source-side and target-side architectural concepts}
\label{tab:traceability}
\small
\begin{tabular}{@{}p{0.27\textwidth} p{0.29\textwidth} p{0.34\textwidth}@{}}
\toprule
\textbf{Source-side concept} & \textbf{Target-side/web concept} & \textbf{Role in migration} \\
\midrule
.NET/C\# computational model
& TypeScript model/runtime classes
& Supplies the browser execution realization checked against reference outputs \\

\texttt{OrbitalForecastCalculator}
& \texttt{OrbitalForecast}
& Generated model components, formulas, configuration, and connections \\

C\# \texttt{Performer}
& \texttt{OrbitalForecastCalculation}
& Developer-supplied execution wrappers \\

Motion-equation aliases such as \texttt{x}, \texttt{y}, \texttt{z}, \texttt{u}, \texttt{v}, \texttt{w}
& TypeScript alias values with the same semantic roles
& Transfers initial state and exposes calculated variables \\

Data consumer / chart object
& TypeScript data consumer object
& Collects calculated values during execution \\

Measurements
& TypeScript measurement interfaces and maps
& Preserve typed access to model variables \\

\texttt{RungeProcessor} and fixed-step execution
& TypeScript Runge processor and ODE runtime execution
& Advances the model over time \\

Domain-specific motion components
& TypeScript motion and \texttt{Motion6D} components
& Provide target-side implementations of domain behavior \\

ASP.NET Core orbital endpoint
& React client communication layer
& Enables reference execution and client/server comparison \\
\bottomrule
\end{tabular}
\end{table*}

\section{TypeScript Runtime and Web Execution}

The TypeScript runtime connects generated models to their execution environment. Aliases provide named access to state for initialization and retrieval. Measurement interfaces and maps expose component values to formulas and output consumers during execution.

For orbital forecasting, the execution wrapper constructs a \texttt{DataRuntimeConsumerODE} with a \texttt{Motion6DFactory}, which supplies the differential-equation processor. After state initialization, the processor advances the model in fixed steps and the consumer reads selected alias values to collect trajectory points. Table~\ref{tab:traceability} relates these runtime elements to their C\# counterparts.

The React client creates the calculation object, passes the forecast inputs, and displays the resulting trajectory. The following case study describes the generated graph and its application wrapper.

\section{Case Study: Orbital Forecasting}

\begin{figure*}[t]
\centering
\includegraphics[width=0.66\textwidth]{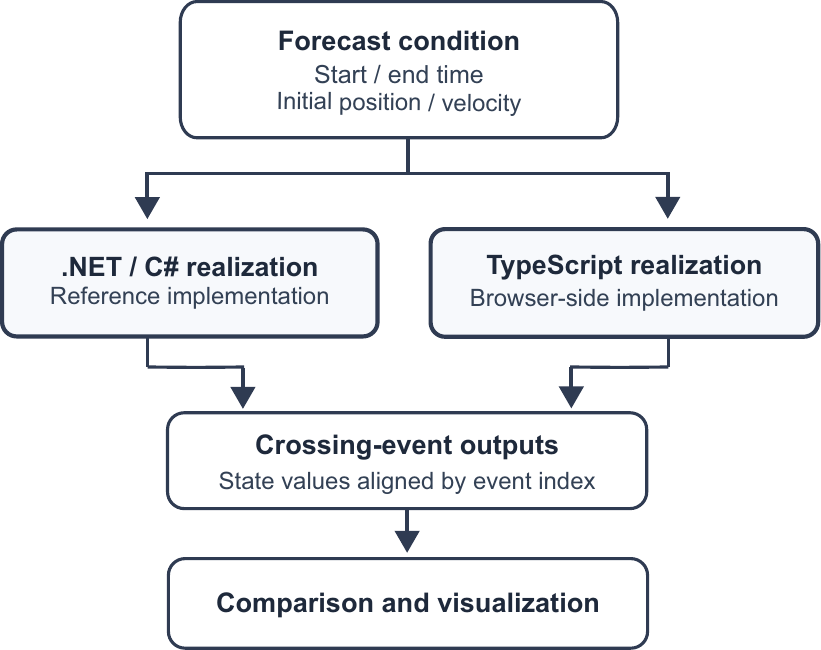}
\caption{Orbital forecasting workflow used in the case study. The same forecast setup is executed through the C\# server reference path and the browser-side TypeScript realization; the resulting trajectory points are then used for visualization and reference-based comparison}
\label{figure_orbital_workflow}
\end{figure*}

The orbital forecasting model combines six state variables, motion equations, numerical integration, and trajectory collection. It brings formula components and domain-specific motion components together in a continuous-time calculation.

The generated TypeScript graph, \texttt{OrbitalForecast}, contains eight component subclasses and eleven connection subclasses. These encode the model construction, links, aliases, atmosphere parameters, gravity coefficients, formulas, and crossing condition. The corresponding C\# graph is \texttt{OrbitalForecastCalculator}. The handwritten \texttt{OrbitalForecastCalculation} wrapper extends the TypeScript graph, assigns forecast inputs, selects the numerical runtime and sampling rule, handles cancellation, and collects output records. Reusable runtime libraries implement RK4 integration and the atmosphere and gravity algorithms. The evaluation starts from the distributed C\# and TypeScript graph implementations and their execution wrappers.

A forecast condition specifies the start and end times and the initial position and velocity. Recorded outputs contain time, position, and velocity; Table~\ref{tab:orbital-notation} defines their notation.

The orbital model is a six-dimensional ordinary differential equation in which the position derivatives are the velocity components,
\[
\dot{x} = u, \quad \dot{y} = v, \quad \dot{z} = w,
\]
and the acceleration components are provided by the motion-equation component of the migrated model. Position is expressed in kilometers and velocity in kilometers per second. Both implementations use the classical fourth-order Runge--Kutta method with a fixed one-second step. Outputs are collected when \texttt{Recursive.y} becomes true: the current \(z\) value is positive and the stored preceding value is negative. Thus the recorded points are discrete upward crossings of the \(z=0\) plane, rather than every integration step. No interpolation to the exact crossing is performed.

\begin{table}[t]
\centering
\caption{State-variable notation for the orbital forecasting case study}
\label{tab:orbital-notation}
\small
\begin{tabular}{@{}p{0.22\linewidth} p{0.64\linewidth}@{}}
\toprule
\textbf{Notation} & \textbf{Role} \\
\midrule
\(x,y,z\) & Cartesian position aliases in the shared coordinate frame \\
\(u,v,w\) & Cartesian velocity aliases in the same frame \\
\(\mathbf{p}_i\) & Position vector with components \texttt{x}, \texttt{y}, and \texttt{z} at trajectory point \(i\) \\
\(\mathbf{v}_i\) & Velocity vector with components \texttt{u}, \texttt{v}, and \texttt{w} at trajectory point \(i\) \\
\(C\#\), \(TS\) & Superscripts denoting the server-side C\# and client-side TypeScript execution paths \\
\bottomrule
\end{tabular}
\end{table}

Figure~\ref{figure_orbital_workflow} shows the comparison workflow.

\section{Evaluation}

\subsection{Experimental protocol and scenarios}

We compare the C\# server implementation with the TypeScript implementation executed in headless Google Chrome on Windows x64. The server uses .NET in Debug configuration, and Node.js orchestrates the browser calls. Online Resource~1 specifies the execution environment; the \path{experiments/orbital/replay} directory in the repository contains the source manifests, inputs, outputs, and reproduction instructions.

Both implementations receive the same numeric requests for the four initial conditions in Table~\ref{tab:orbital-scenarios}. Times are OLE Automation seconds, with the displayed dates interpreted as UTC. Each scenario covers 538404 seconds with a one-second RK4 step and includes the final grid point. The input requests, expected event counts, and comparison tolerances were fixed in \path{protocol.json} before execution.
\begin{table*}[!b]
\caption{Initial conditions supplied to both implementations in the four orbital scenarios. Position is in km and velocity in km/s. Scenarios 2--4 change the initial position or vertical velocity.}
\label{tab:orbital-scenarios}
\centering
\small
\begin{tabular}{lllllll}
\hline
Run & $x$ & $y$ & $z$ & $v_x$ & $v_y$ & $v_z$ \\
\hline
1 & -5448.34815324 & -4463.93698421 & 0 & -0.985394777432 & 1.21681893834 & 7.45047785592 \\
2 & -5458.34815324 & -4463.93698421 & 0 & -0.985394777432 & 1.21681893834 & 7.45047785592 \\
3 & -5458.34815324 & -4453.93698421 & 0 & -0.985394777432 & 1.21681893834 & 7.45047785592 \\
4 & -5458.34815324 & -4453.93698421 & 0 & -0.985394777432 & 1.21681893834 & 7.45057785592 \\
\hline
\multicolumn{7}{l}{Start (UTC): 2026-06-24T03:45:06.449; end (UTC): 2026-06-30T09:18:30.449.}
\end{tabular}
\end{table*}

\subsection{Temporal sampling and scheduler validation}

At each grid point after the initial time, the runtime advances the state over the preceding interval, evaluates the crossing condition, and collects the selected outputs. The state and its timestamp therefore refer to the same grid point. The interval contains 538405 grid observations and 538404 integration steps.

Twenty analytic checks test the scheduler with constant and time-dependent derivatives, nonzero initial times, fractional steps, conditional output, cancellation, and interval endpoints. All 20 pass. These checks use analytically integrated mock derivatives to assess scheduling separately from RK4's numerical approximation.

\subsection{Comparison criteria and numerical scale}

We pair outputs by upward-crossing event index and require identical timestamps before assessing state agreement. For each pair, the position and velocity residuals are
\[
\begin{aligned}
d_p(i)&=\lVert\mathbf{p}^{\mathrm{C\#}}_i-\mathbf{p}^{\mathrm{TS}}_i\rVert_2,\\
d_v(i)&=\lVert\mathbf{v}^{\mathrm{C\#}}_i-\mathbf{v}^{\mathrm{TS}}_i\rVert_2.
\end{aligned}
\]
The regression tolerances are $10^{-6}$~km (1~mm) and $10^{-9}$~km/s (1~$\mu$m/s). At characteristic scales of 7000~km and 7.5~km/s, these correspond to relative differences of approximately $1.4\times10^{-10}$ and $1.3\times10^{-10}$. These engineering thresholds specify the agreement required of the two implementations.

Both implementations use binary64 arithmetic, with unit roundoff $u_{\mathrm{fp}}=2^{-53}$~\cite{higham2002accuracy}. For $N=538404$ steps, the dimensional reference scale $Nu_{\mathrm{fp}}\simeq5.98\times10^{-11}$ corresponds to $4.19\times10^{-7}$~km and $4.48\times10^{-10}$~km/s at the stated position and velocity scales. This supplies numerical context for the tolerances, not a bound on trajectory error or an explanation of the observed residuals: each integration step contains many operations, and the dynamics can amplify perturbations. The comparison measures agreement between two RK4 implementations; the full residual histories show how that agreement evolves during integration.

\subsection{Results and diagnostic controls}

\begin{figure*}[t]
\centering
\includegraphics[width=\textwidth]{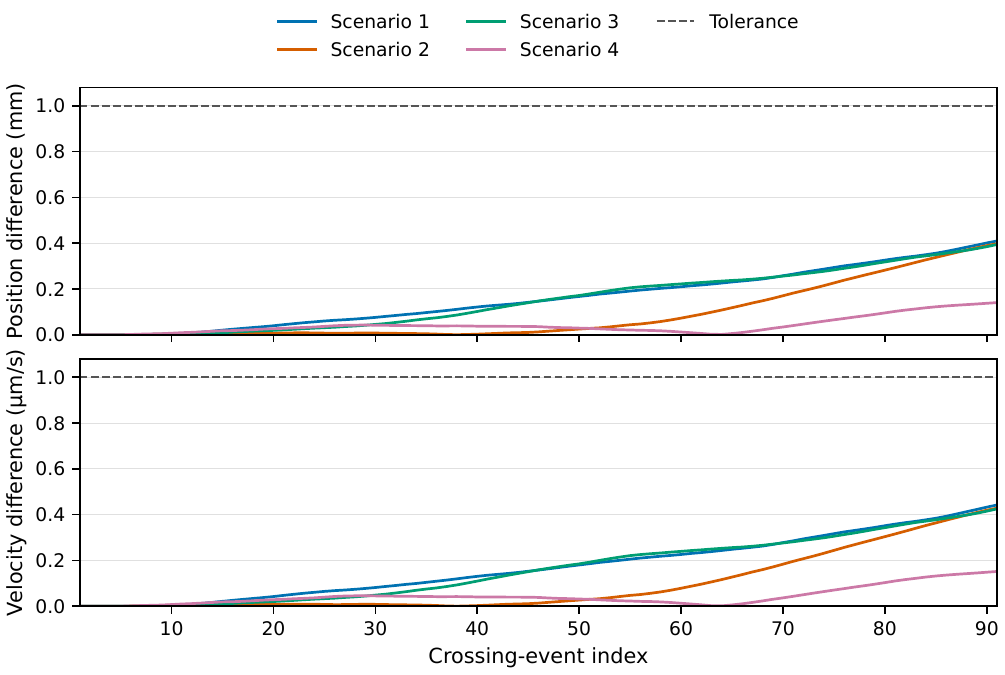}
\caption{Position and velocity residuals against crossing-event index for the four scenarios. Dashed lines show the 1~mm and 1~$\mu$m/s tolerances; paired timestamps agree exactly}
\label{fig:orbital-residuals}
\end{figure*}

All four scenarios produced 91 events per path: 364 pairs with exactly equal numeric timestamps. Table~\ref{tab:orbital-validation-results} reports the measured residuals. All satisfy the specified criteria. Figure~\ref{fig:orbital-residuals} shows their evolution over the event sequence, including accumulation during the long integration.

\begin{table*}[t]
\caption{C\#/browser TypeScript comparison for the four orbital scenarios. Paired event timestamps agree exactly. Position and velocity residuals are Euclidean norms in km and km/s}
\label{tab:orbital-validation-results}
\centering\small
\begin{tabular}{llllll}
\hline
Run & Events & Max. pos. diff. & Mean pos. diff. & Max. vel. diff. & Mean vel. diff. \\
\hline
1 & 91 & \(4.10 \times 10^{-7}\) & \(1.60 \times 10^{-7}\) & \(4.42 \times 10^{-10}\) & \(1.72 \times 10^{-10}\) \\
2 & 91 & \(3.97 \times 10^{-7}\) & \(9.00 \times 10^{-8}\) & \(4.28 \times 10^{-10}\) & \(9.70 \times 10^{-11}\) \\
3 & 91 & \(3.94 \times 10^{-7}\) & \(1.52 \times 10^{-7}\) & \(4.25 \times 10^{-10}\) & \(1.64 \times 10^{-10}\) \\
4 & 91 & \(1.41 \times 10^{-7}\) & \(4.05 \times 10^{-8}\) & \(1.52 \times 10^{-10}\) & \(4.38 \times 10^{-11}\) \\
\hline
\end{tabular}
\end{table*}

The comparator detects all three prescribed negative controls: adding 2~mm to one position component, 2~$\mu$m/s to one velocity component, or one second to one timestamp makes the comparison fail. The repository package supplies the raw records and scripts used to calculate every table entry and plotted residual.

\section{Case Study: Trading Strategy}

\begin{figure*}[t]
\centering
\includegraphics[width=0.70\textwidth]{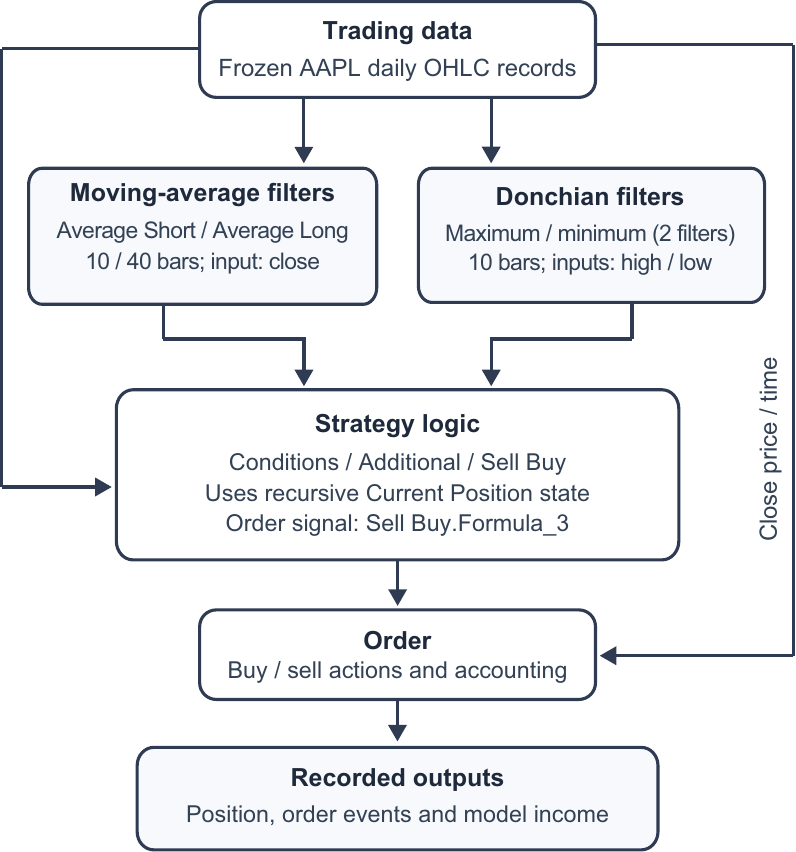}
\caption{Simplified component structure of the trading strategy case study. Historical trading data is processed by average and Donchian Channel filters; formula and recursive-position components compute Sell Buy.Formula\_3, which is used by the order component to generate trading actions}
\label{trading_strategy_schematic}
\end{figure*}

The trading strategy provides a second case with historical input, moving-window filters, recursive state, and order generation.

The strategy is organized as a component graph. The \texttt{Trading} plugin provides historical market data, including open-high-low-close values and time. The \texttt{Order} plugin represents buy and sell actions. For the evaluation, both realizations used the same daily AAPL input data for the interval from January 1, 2022, inclusive, to January 1, 2024, exclusive. The resulting dataset contains 501 trading-day records from January 3, 2022, through December 29, 2023. The strategy uses 10- and 40-day moving averages and 10-day Donchian high and low windows. The order component uses \texttt{Sell Buy.Formula\_3} as the position signal, and the close price is used as the buy and sell price.

\begin{table*}[t]
\caption{Behavioral validation results for the trading strategy case. The C\# and TypeScript realizations use the same daily AAPL input and are aligned by trading date}
\label{tab:trading-validation}
\centering
\small
\begin{tabular*}{\textwidth}{@{\extracolsep{\fill}}lccc@{}}
\hline
Measure & C\# & TypeScript & Difference \\
\hline
\multicolumn{4}{l}{\textit{Input scenario}} \\
Symbol & AAPL & AAPL & -- \\
Bar interval & 1 day & 1 day & -- \\
Historical interval & \multicolumn{2}{c}{2022-01-01 to 2024-01-01 (end exclusive)} & -- \\
Aligned daily records & 501 & 501 & 0 \\
Short / long moving-average windows & 10 / 40 & 10 / 40 & -- \\
Donchian high / low windows & 10 / 10 & 10 / 10 & -- \\
\hline
\multicolumn{4}{l}{\textit{Behavioral preservation}} \\
Buy events & 231 & 231 & 0 \\
Sell events & 231 & 231 & 0 \\
Position mismatches & -- & -- & 0 \\
Order-event mismatches & -- & -- & 0 \\
Records with any mismatch & -- & -- & 0 \\
\hline
\multicolumn{4}{l}{\textit{Numerical agreement}} \\
Max. abs. error, short moving average & -- & -- & 0 \\
Max. abs. error, long moving average & -- & -- & 0 \\
Max. abs. error, Donchian high & -- & -- & 0 \\
Max. abs. error, Donchian low & -- & -- & 0 \\
Max. abs. error, buy/sell prices & -- & -- & 0 \\
Max. abs. error, accumulated income & -- & -- & 0 \\
Final model income & -4.2099533081 & -4.2099533081 & 0 \\
\hline
\end{tabular*}
\end{table*}

The model contains two moving-average filters, two Donchian Channel filters, formula components, and a recursive current-position element. Figure~\ref{trading_strategy_schematic} summarizes their connections. The position signal depends on both calculated conditions and stored position state, so conversion must retain the component connections and state handling as well as the formulas.

The experiment uses the generated \texttt{DonchianDesktop} models from \texttt{AspireTradingApp}. The TypeScript runner bundles the model with esbuild and executes it in Node.js. Both runners read the same frozen daily-data file. Indicator windows include the current bar, and the initial recursive state is zero. Both runners consume the complete daily sequence, including its first record, without resampling the aggregated bars.

Outputs are aligned by trading date. The comparison covers the moving-average and Donchian values, position states, buy and sell events, event prices, and accumulated model income.

Numerical values are compared using an absolute tolerance of \(10^{-10}\) and a relative tolerance of \(10^{-12}\). For numerical values \(a\) and \(b\), acceptance requires
\[
|a-b| \leq 10^{-10} + 10^{-12}\max(|a|,|b|).
\]
Position states and order events are compared exactly. Position codes are 0 (flat), 1 (short), and 2 (long). Events are read from the order component's buy and sell price measurements. Records produced during the initial warm-up period are retained in the aligned sequence, with unavailable moving-window values represented consistently as \texttt{null} values. Accumulated income is the output of the model's order-accounting component and is compared as a state variable.

Both realizations processed 501 aligned daily records and generated 231 buy events and 231 sell events (Table~\ref{tab:trading-validation}). No record mismatches were observed, and all recorded numerical differences were zero, including the final accumulated model income; final position states also matched.

\section{Related Work}

Source-to-source translators include tools for converting Java or C\# programs to JavaScript or TypeScript and programmable transpilation approaches~\cite{jsweet,bridge,duoglot}. The Simplified Wrapper and Interface Generator (SWIG) instead connects existing native components to other languages through generated wrappers~\cite{beazley_swig}. \toolname{} operates on the component model and generates its target configuration and calculations using supported runtime components.

Blazor WebAssembly executes C\# and .NET code in the browser through a WebAssembly-based runtime~\cite{blazor}. The converter produces TypeScript code, allowing the model to integrate with TypeScript libraries and application logic.

CrossEcore generates code from Ecore models with embedded OCL and supplies target runtime APIs, including a TypeScript implementation~\cite{schwichtenberg2018crossecore}. It shares the architectural separation between model-specific generated code and reusable runtime support. Its input is a class-and-constraint model; \toolname{} maps executable computational components, including formula trees, aliases, measurement bindings, and numerical processors. The distinction lies in these mappings and the execution behavior they support.

SDEverywhere translates Vensim and Stella system-dynamics models into JavaScript or C, supports WebAssembly deployment, and provides model-comparison tools~\cite{sdeverywhere}. OpenModelica's Emscripten backend compiles simulation code to JavaScript~\cite{short2014openmodelicajs}. Modiator translates a supported Modelica subset to JavaScript and uses Sundials solvers compiled to WebAssembly~\cite{elmqvist2025modiator}. These systems provide related routes from executable models to browser applications. \toolname{} addresses .NET computational component graphs through explicit TypeScript component mappings and evaluates their execution in continuous and discrete case studies.

Model-driven engineering provides methods for generating software from models, including modular model transformations~\cite{mde,hemel2010codegen}. Sirius Web, WebGME, and Graphical Language Server Platform (GLSP)-based environments support web modeling interfaces, language workbenches, and collaborative tooling~\cite{giraudet_sirius_2024,webgme,bork_vision_2023}. The GLSP reference architecture documents reusable infrastructure and development experience~\cite{metin_bork_2025}. These tools primarily support creating and managing models; \toolname{} focuses on generating an executable TypeScript realization of an existing computational model.

Concrete-syntax transformation systems such as the Concrete Grammar Transformation Language (CGTL) specify translations over parsed language structures~\cite{lano2024concrete}, while example-based approaches learn code-generation rules from paired source and target examples~\cite{lano2023cgbe}. Generative process-aware information systems also separate model-derived artifacts from reusable runtime services and handwritten additions~\cite{drave2022pais}. The present approach uses developer-supplied mappings of component objects and their interfaces as its generation rules.

\section{Discussion and Limitations}
\label{sec:limitations}

The two case studies connect different computational behaviors to the same component architecture. Orbital forecasting combines numerical integration with event-based output, while the trading model combines moving-window calculations, recursive state, and orders. Their validation depends on explicit alignment rules: equal grid times for orbital events and equal dates and input sequences for trading records. These rules make component behavior observable across the language boundary.

The empirical scope is four orbital configurations and one trading configuration using a single symbol and daily dataset. The orbital evaluation starts from the distributed generated graphs; the corresponding editor-model file is not available among the identified artifacts for an end-to-end regeneration check. Browser execution is evaluated for the orbital case, while the recorded trading comparison uses Node.js. The results establish output agreement for these implementations and scenarios. Physical prediction accuracy and trading profitability are separate validation objectives.

The orbital configuration also limits sensitivity to some domain components. The atmosphere component executes, but its drag contribution is numerically negligible: omitting the sampled drag terms leaves the binary64 accelerations unchanged. Both language implementations bind two velocity inputs to the same vector slot. Agreement therefore does not independently validate this shared model configuration. The reported residuals characterize the documented runtime environment and Debug server build; their exact values may vary across platforms and build configurations.

\section{Conclusion}

\toolname{} maps BP\_Simulator computational models to TypeScript through explicit rules for components, connections, aliases, and formulas. Reusable libraries supply the execution behavior required by these models, including numerical integration, data access, and state handling. This organization makes the boundary between generated model code and developer-provided runtime support explicit.

Execution of the distributed model implementations demonstrates agreement across the C\#/TypeScript boundary in four orbital scenarios and a recorded trading comparison. All 364 orbital event pairs have equal timestamps and state residuals within the specified tolerances; the 501 trading records have matching numerical outputs and order events. The component mappings, execution protocol, and reproducibility materials provide a concrete basis for deploying supported computational models in TypeScript applications.

\section*{Supplementary Information}

\textbf{Online Resource 1 (ESM\_1.pdf).} Orbital experimental protocol, execution environment, sampling rules, and diagnostic checks. The same notes are included as an appendix to this manuscript.

\section*{Statements and Declarations}

\subsection*{Funding}
The authors did not receive support from any organization for the submitted work.

\subsection*{Ethics Approval}
The study did not involve human participants, human-subject data, or animal subjects. Ethics approval and informed consent were not required.

\subsection*{Author Contributions}

\textbf{PI:} Conceptualization; Methodology; Software; Data Curation; Investigation; Formal Analysis; Writing -- Original Draft; Writing -- Review \& Editing; Visualization; Validation.

\textbf{MS:} Methodology; Software; Data Curation; Investigation; Formal Analysis; Writing -- Original Draft; Writing -- Review \& Editing; Visualization; Project Administration; Validation.

\subsection*{Competing interests}
The authors have no competing interests to declare that are relevant to the content of this article.

\subsection*{Availability of Data and Material}
The orbital input protocol, raw outputs, comparison and plotting scripts, analytic scheduler checks, and source manifests are available in the repository under \href{https://github.com/SynthyLink/bp_simulator_converter/tree/6a234ca788092841edf46fd50368273b10a188f7/experiments/orbital/replay}{\nolinkurl{experiments/orbital/replay}}. The trading dataset, outputs, and execution instructions are available under \path{experiments/trading} in the repository. These materials support reproduction of the reported comparisons.

\subsection*{Code Availability}
The converter, runtime libraries, example applications, and orbital reproduction package are available under the MIT license at
\url{https://github.com/SynthyLink/bp_simulator_converter},
\href{https://github.com/SynthyLink/bp_simulator_converter/tree/6a234ca788092841edf46fd50368273b10a188f7}{commit \nolinkurl{6a234ca788092841edf46fd50368273b10a188f7}}.
The source used for the trading comparison is identified by
\href{https://github.com/SynthyLink/bp_simulator_converter/tree/f40af51ce5e2629a9e26b89cdf036a864209f0b8}{commit \nolinkurl{f40af51ce5e2629a9e26b89cdf036a864209f0b8}}.
The orbital package specifies the exact execution sources through its protocol, source hashes, and checkout instructions. A worked conversion procedure for the FeedBackFormula model is included in the repository.

\setlength{\bibsep}{0.1em}
\bibliography{backend-frontend-converter}
\clearpage
\onecolumn
\setcounter{table}{0}
\renewcommand{\thetable}{S\arabic{table}}
\section*{Online Resource 1: Reproducibility Notes for the Orbital Forecasting Case}

These notes describe the orbital experimental protocol and the reproduction package in \href{https://github.com/SynthyLink/bp_simulator_converter/tree/6a234ca788092841edf46fd50368273b10a188f7/experiments/orbital/replay}{\nolinkurl{experiments/orbital/replay}} at repository commit \nolinkurl{6a234ca7}. The package README gives the exact source checkout, build, and execution commands; source hashes identify the implementation used for each recorded run.

\supplementsection{Execution environment}

The orbital application is under \path{examples/orbital/AspireOnlineConverter} in the MIT-licensed repository identified in the Code Availability statement. Its \path{AspireOnlineConverter.Server} project provides the C\# implementation, and its \path{frontend} directory provides the TypeScript application. The TypeScript calculation executes in headless Google Chrome; Node.js and Playwright orchestrate the calls.

The experiment used Windows x64 (10.0.26200), .NET SDK 10.0.301 and runtime 10.0.9, Node.js 24.12.0, npm 11.6.2, TypeScript 5.9.3, Vite 8.0.16, Google Chrome 154.0.8037.98, and Playwright 1.62.1. The server was built in Debug configuration, and frontend dependencies were installed from the package lockfile. Execution records are dated 13 October 2026 UTC. Two browser executions of all four scenarios produced identical timestamps and state values.

\supplementsection{Input protocol and sampling}

\path{evidence/protocol.json} specifies the four numeric input requests, one-second step, expected event counts, position and velocity tolerances, and exact timestamp criterion. The protocol was fixed before execution. The 538404-second interval runs from 2026-06-24T03:45:06.449 to 2026-06-30T09:18:30.449, with both date values interpreted as UTC. The corresponding OLE Automation times are 3991434306.449 and 3991972710.449 seconds. Both implementations receive the same numeric requests directly.

Both calculations use classical fourth-order Runge--Kutta integration. The runtime advances the state to each grid time before evaluating the output condition. The interval contains 538405 grid observations and 538404 integration steps. Output is selected when the current $z$ is positive and its preceding value was negative. These discrete upward crossings exclude the initial point at zero and use no interpolation. The comparison pairs records by event index and tests their timestamps for exact numeric equality.

\supplementsection{Reproduction and analysis}

The repository package contains the execution runner \path{replay-portable.cjs}, the input protocol, environment records, source manifests, and raw C\# and browser outputs. Its README specifies how to reconstruct the recorded execution sources and start the two local services. Browser calls import the application's \texttt{OrbitalForecast\allowbreak{}Calculation} class through Vite. The source manifests and \path{evidence/execution-provenance.json} document the correspondence between the executable implementation and the supplied records.

\begin{samepage}
Run the following commands from \path{experiments/orbital/replay} to recompute the numerical summaries and residual plot from the recorded outputs:
\begin{verbatim}
node analyze-replay.cjs evidence
python plot-residuals.py evidence
\end{verbatim}
\end{samepage}
The analysis writes \path{comparison.json} and \path{event-residuals.csv}. The plotting script uses the latter to generate the residual figure. The README also supplies the command for running \path{scheduler-regression.cjs} against the reconstructed source. \path{SHA256SUMS.json} records the package's file hashes.

\supplementsection{Scheduler and comparator checks}

Twenty analytic checks cover constant and time-dependent derivatives, nonzero starting times, fractional steps, conditional output, cancellation, and interval endpoints. Analytically integrated mock derivatives isolate scheduling from the numerical approximation of RK4. All 20 checks pass for the evaluated scheduler.

The numerical comparator checks finite values, matching inputs, event counts, timestamps, and Euclidean position and velocity residuals. All four scenarios satisfy the 1~mm and 1~$\mu$m/s thresholds, with equal timestamps across 364 paired events. Three separate perturbations assess comparator sensitivity: a 2~mm change to one position component, a 2~$\mu$m/s change to one velocity component, and a one-second change to one timestamp. Each perturbation causes the relevant comparison to fail.

\supplementsection{Atmospheric sensitivity}

The diagnostic \path{evidence/atmosphere-clock-probe.json} records intermediate values from the orbital graph. Atmospheric density is near $2.07\times10^{-15}$, the coefficient $s$ is $1.62\times10^{-13}$, and the initial drag acceleration terms are at most $1.93\times10^{-26}$~km/s$^2$. Removing these terms leaves the sampled accelerations unchanged in binary64. This quantifies the sensitivity limitation discussed in Section~10 of the manuscript.

\end{document}